\documentclass[10pt,twocolumn,letterpaper]{article}

\usepackage[]{}      % To produce the REVIEW version
\usepackage{cvpr}              % To produce the CAMERA-READY version
\usepackage[accsupp]{axessibility}  % Improves PDF readability for those with disabilities.
\usepackage{graphicx}
\usepackage{amsmath}
\usepackage{amssymb}
\usepackage{booktabs}
\usepackage{titling}
\usepackage[table]{xcolor}
\usepackage{enumitem}
\usepackage{placeins}

\usepackage[pagebackref,breaklinks,colorlinks]{hyperref}

\usepackage[capitalize]{cleveref}
\crefname{section}{Sec.}{Secs.}
\Crefname{section}{Section}{Sections}
\Crefname{table}{Table}{Tables}
\crefname{table}{Tab.}{Tabs.}

\def\cvprPaperID{26} % *** Enter the CVPR Paper ID here
\def\confName{CVPR}
\def\confYear{2022}

\begin{document}

\title{Segmenting across places: The need for fair transfer learning with satellite imagery}

\author{Anonymous CVPR submission \\
{PaperID: 26}
}
\author{Miao Zhang \qquad Harvineet Singh \qquad Lazarus Chok \qquad Rumi Chunara \\
New York University\\
{\tt\small \{miaozhng, hs3673, lcc9673, rumi.chunara\}@nyu.edu}
% \and
% Harvineet Singh \\
% New York University \\
% {\tt\small hs3673@nyu.edu}
% \and
% Lazarus Chok \\
% New York University \\
% {\tt\small lazaruschokch@gmail.com}
% \and
% Rumi Chunara\\
% New York University \\
% {\tt\small rumi.chunara@nyu.edu}
}

\date{}
\maketitle

%%%%%%%%% ABSTRACT
\begin{abstract}
The increasing availability of high-resolution satellite imagery has enabled the use of machine learning to support land-cover measurement and inform policy-making. However, labelling satellite images is expensive and is available for only some locations. This prompts the use of transfer learning to adapt models from data-rich locations to others. Given the potential for high-impact applications of satellite imagery across geographies, a systematic assessment of transfer learning implications is warranted. In this work, we consider the task of land-cover segmentation and study the fairness implications of transferring models across locations. We leverage a large satellite image segmentation benchmark with 5987 images from 18 districts (9 urban and 9 rural). Via fairness metrics we quantify disparities in model performance along two axes -- across urban-rural locations and across land-cover classes. Findings show that state-of-the-art models have better overall accuracy in rural areas compared to urban areas, through unsupervised domain adaptation methods transfer learning better to urban versus rural areas and enlarge fairness gaps. In analysis of reasons for these findings, we show that raw satellite images are overall more dissimilar between source and target districts for rural than for urban locations. This work highlights the need to conduct fairness analysis for satellite imagery segmentation models and motivates the development of methods for fair transfer learning in order not to introduce disparities between places, particularly urban and rural locations. 
 
% Given the rising concern of detrimental social impacts caused from predictive model's inequality amongst subgroups, here we integrate fairness analysis to semantic segmentation in transfer learning settings. We implement two types of transfer: across geographical locations and across rural and urban areas on a real-world satellite image dataset, and evaluate using proposed fairness metrics. This work is an extension of existing fairness studies by investigating fairness change of satellite image segmentation models in a transfer learning framework.

\end{abstract}

%%%%%%%%% BODY TEXT
\section{Introduction}
\label{sec:intro}
% Advanced remote sensing technologies provide high-resolution satellite imagery of Earth's surface, which enables the training of machine learning models for various image interpretation tasks. 
Satellite imagery is becoming readily available with around 1030 active satellites that are dedicated to earth observation \cite{ucsdata}. Out of the different spectra of imagery available from such satellites, visible spectrum imagery is particularly relevant for many applications based on the extremely high resolution and according ability to resolve specific objects of interest ~\cite{carleer2005assessment}. Consequently, satellite images combined with semantic segmentation, the task of clustering parts of an image together which belong to the same object class, can be used to detect objects ranging from natural features (water bodies, forests) to human land-use types (buildings, roads). The extracted information is being applied in a wide range of settings including urban planning ~\cite{moghalles2021multi}, modelling disease spread ~\cite{abdur2019deep}, aiding disaster relief efforts ~\cite{gupta2019cnn, zhao2020building}, and detecting and mapping environmental phenomena ~\cite{krestenitis2019oil, wu2020deep}. However, because segmentation models employ supervised learning, availability of ground truth data is a major bottleneck for their training. Annotation for the segmentation task is particularly labor intensive as it requires fine-grained labels at the level of pixels which results in incomplete or noisy ground truth data ~\cite{schmitt2020weakly}. In such situations, generalizing existing models to non-annotated data by \textit{transfer learning} is a widely applied solution ~\cite{torrey2010transfer, rolf2021generalizable}.

Transfer learning uses knowledge learnt from the same or related tasks to improve learning on the task at hand (see Pan and Yang~\cite{pan2010survey} for a survey). We will focus on a type of transfer learning setting called \textit{domain adaptation}, where we have a single task but the train and test domains may differ. The key challenge here is the discrepancy in data distributions between domains. In the case of satellite imagery, the discrepancies commonly result from transferring models to new geographies where the landscapes are dissimilar to where the model was trained. For example, Islam ~\cite{islam2021deep} finds that a well-trained seagrass detection model from satellite images fails when tested at other locations with different seagrass density. To mitigate the degrading effects of domain discrepancies on segmentation accuracy, previous work has re-designed network architectures ~\cite{li2007semantic}, loss functions ~\cite{henry2018road, volpi2015semantic}, and batch normalization methods ~\cite{ortiz2020local} to improve model generalization. Other approaches include using labels at a coarser granularity for the target domain (e.g. image-level labels) as weak supervision~\cite{iqbal2020weakly} and learning latent representations shared between source and target domains to help in adaptation~\cite{liu2018hyperspectral,li2019transfer}.

Simultaneously, while machine learning approaches have been used to improve prediction in a variety of tasks, recent studies have highlighted concerns towards model fairness, exhibited by performance disparities across sensitive groups based on geography, demographics, and economic indicators ~\cite{mcgovern2021need, zhou2021radfusion}. A push for model fairness aligns with the ideal of equity defined by World Health Organization as ``Equity is the absence of unfair, avoidable or remediable differences among groups of people, whether those groups are defined socially, economically, demographically, or geographically or by other dimensions of inequality (e.g. sex, gender, ethnicity, disability, or sexual orientation).''  
Real-world examples have demonstrated the harmful effects of unfair machine learning models, such as facial recognition software that performs worse on darker women \cite{buolamwini2018shades} and advertisement systems that deliver economic opportunity-related ads less often to women than men \cite{lambrecht2019ads}.
% due to gender-imbalance.
Indeed, discriminatory issues persist even in state-of-the-art learning methods ~\cite{mehrabi2021survey}. Expanding types of data used in machine learning tasks, such as satellite imagery, enables increased use in a wide range of daily-life applications and ever-increasing social impacts. Accordingly, broader aspects and viewpoints of performance, such as fairness, need to be ascertained in multiple machine learning subareas.

In this work we study \textbf{the fairness impacts of transfer learning with satellite imagery}. To accomplish this goal, we test multiple semantic segmentation models across different geographies. We then assess if such models made fair predictions on both the source and the target data. We focus on differences between urban and rural areas (i.e. urban/rural categorization is the sensitive attribute) due to persistent and striking disparities between urban and rural areas, especially for poor populations ~\cite{brahmadesamfairly, bakker2020fair}. The unfairness criterion in this work is based on differences in error rates across protected groups where the error rate is computed using Intersection-over-Union (IoU), a standard segmentation metric. We also examine model performance disparity across different land-cover classes. Results show that existing domain adaptation methods do not maintain fairness properties on transfer, either across protected groups or feature classes. This work serves as a valuable demonstration of fairness being an critical issue in transfer learning using a large freely-available satellite imagery dataset.

Important takeaways are as follows.
\begin{itemize}[leftmargin=*]
\item Studied models have better overall accuracy (via mean IoU over the 7 classes) on rural districts as compared to urban districts.

\item For common unsupervised domain adaptation methods, transfer accuracy is improved, but at the cost of fairness; the performance gap between rural and urban group is enlarged indicating the need to design new methods that transfer well for both the groups.

% *Feature-wise performance differs

\item Investigating reasons for the above findings, we find that images from rural districts differ more across locations than those of urban districts.
% We investigate reasons for the findings (via rural/urban image differences/class distributions, etc.)
\end{itemize}

\section{Related Work}
Before discussing prior work on transfer learning for satellite images, we describe some of the alternative ways to address label scarcity and their shortcomings. Lastly, we summarize work in the nascent area of fair transfer learning.

\textbf{Approaches to tackle annotation burden for satellite images}
% Given the increasing amount and utility of visible spectrum satellite imagery, yet simultaneous
Given the difficulty in labelling data for semantic segmentation of satellite images, Schmitt, \textit{et al.} ~\cite{schmitt2020weakly} developed weakly-supervised learning methods, where noisy, limited, or imprecise data sources are used to provide supervision signal. Previous work has leveraged the spatial context to develop unsupervised losses which, for example, penalize nearby pixels with different predicted labels ~\cite{volpi2015semantic, obukhov2019gated}. In another approach, Castillo-Navarro, \textit{et al.} ~\cite{castillo2021semi} proposed auxiliary losses based on self-supervised image reconstruction to improve the performance on the main task of image segmentation. To improve efficiency of label-use, Wang \textit{et al.}~\cite{wang2020weakly} transferred classification models trained with image-level labels to image segmentation tasks and achieved high accuracy. While these approaches demonstrate successful combination of labeled and unlabeled images, they assume that the images are from the same domain (or distribution). However, the assumption of a consistent domain is not realistic for problems involving satellite images which are often from different geographies. Thus, such approaches are not straight-forwardly applicable in our setting.
% Some previous work proposed training segmentation models by jointly minimizing cross-entropy loss for labeled pixels and unsupervised conditional random field loss for unlabeled pixels to extract spatial context information, which for example, penalizes nearby similar pixels with different labels assigned ~\cite{volpi2015semantic, obukhov2019gated}. In another approach, Castillo-Navarro \etal~\cite{castillo2021semi} proposed auxiliary losses based on self-supervised image reconstruction as regularization to boost the performance of the main task, unsupervised image segmentation, performed by a relaxed k-means loss.

\textbf{Transfer learning for satellite images}
Transferability of satellite image segmentation models across different geographic locations has been studied in Ghorbanzadeh \textit{et al.}~\cite{ghorbanzadeh2021comprehensive}. Using train and test sets across 3 different geographies (Taiwan, China, and Japan), they show consistent decrease in evaluation scores upon transfer. Previous work has incorporated domain adaptation methods to deal with the challenge. For instance, Tran \textit{et al.}~\cite{tran2020pp} proposed a two-stage transfer learning structure which generated pseudo-ground truth segmentation labels for target data. Algorithms to improve the quality of such pseudo labels were studied in ~\cite{zou2018unsupervised, mei2020instance}. Data augmentation is another strategy for domain adaptation. Ji \textit{et al.} augments images to simulate perturbations due to atmospheric radiation and demonstrate improved generalization of CNN-based models ~\cite{ji2019scale}. These studies show the promise of adapting models to data from different locations. But, the transfer is only evaluated based on overall accuracy for the domain, such as using Intersection-over-Union (IoU) to measure the overlap between predicted segmentation maps and ground-truth masks. Past work does not study fine-grained measures of model performance on transfer, like how is the performance for different subgroups in the domain (based on sensitive attributes or land-use types) impacted. The risk that discrepancy between domains in transfer learning may impact subgroups unfairly remains unexplored.

\textbf{Fairness in transfer learning}
Following the work in fair machine learning literature \cite{mehrabi2021survey}, we will narrowly classify the study of performance differences between subgroups as \textit{model fairness} analysis. Compared to fairness analysis within the same domain, little work has studied transfer of fair models across domains. The two objectives--improving transfer accuracy and maintaining fairness--can be at odds with each other \cite{xu2021adversarial,singh2021covariate}. Schumann \textit{et al.}~\cite{schumann2019transfer} formalized the problem of fair transfer learning which sets the learning objective to improve accuracy as well as fairness in the target domain. Multiple approaches to fair transfer learning have been proposed~\cite{coston2019fair,oneto2020representations,rezaei2021robust,singh2021covariate,mandal2020ensuring,roh2021sample,li2021federated} that make various assumptions on how the domains differ and what data is available. Even when labels are not available for the target domain, like in our setting, methods typically make the \textit{covariate shift} assumption which says that the labeling rule remains the same between the domains and only the feature distribution changes. In this setting, Coston \textit{et al.}~\cite{coston2019fair} propose a method for fair transfer even in cases where sensitive attributes are absent from one of the domains. Other approaches do not require access to target domain data altogether and instead either make causal assumptions on the discrepancy \cite{singh2021covariate} or hypothesize a set of target domains and optimize against them \cite{mandal2020ensuring,du2021selection}. Finally, Szabó \textit{et al.} \cite{szabo2021tilted} conduct a fairness evaluation of segmentation methods assuming a single domain. 

However, none of the existing works study fair transfer learning for semantic segmentation models. The task differs considerably from the above settings as the input data is high-dimensional, and the model output and loss function for segmentation are different. We take the first step in this direction by demonstrating the need for such methods via a thorough empirical study on a relevant application.
% Lan et al. showed that transfer learning tasks based on geo-linked crime data suffered from discriminatory transfer; target tasks empirically showed lower fairness ~\cite{lan2017discriminatory}. Fairness and transfer studies for labelling of high-dimensional data, such as images, are absent.
% or their application in real-world practice.

\begin{figure}[h!]
  \centering
    \includegraphics[width=8.3cm]{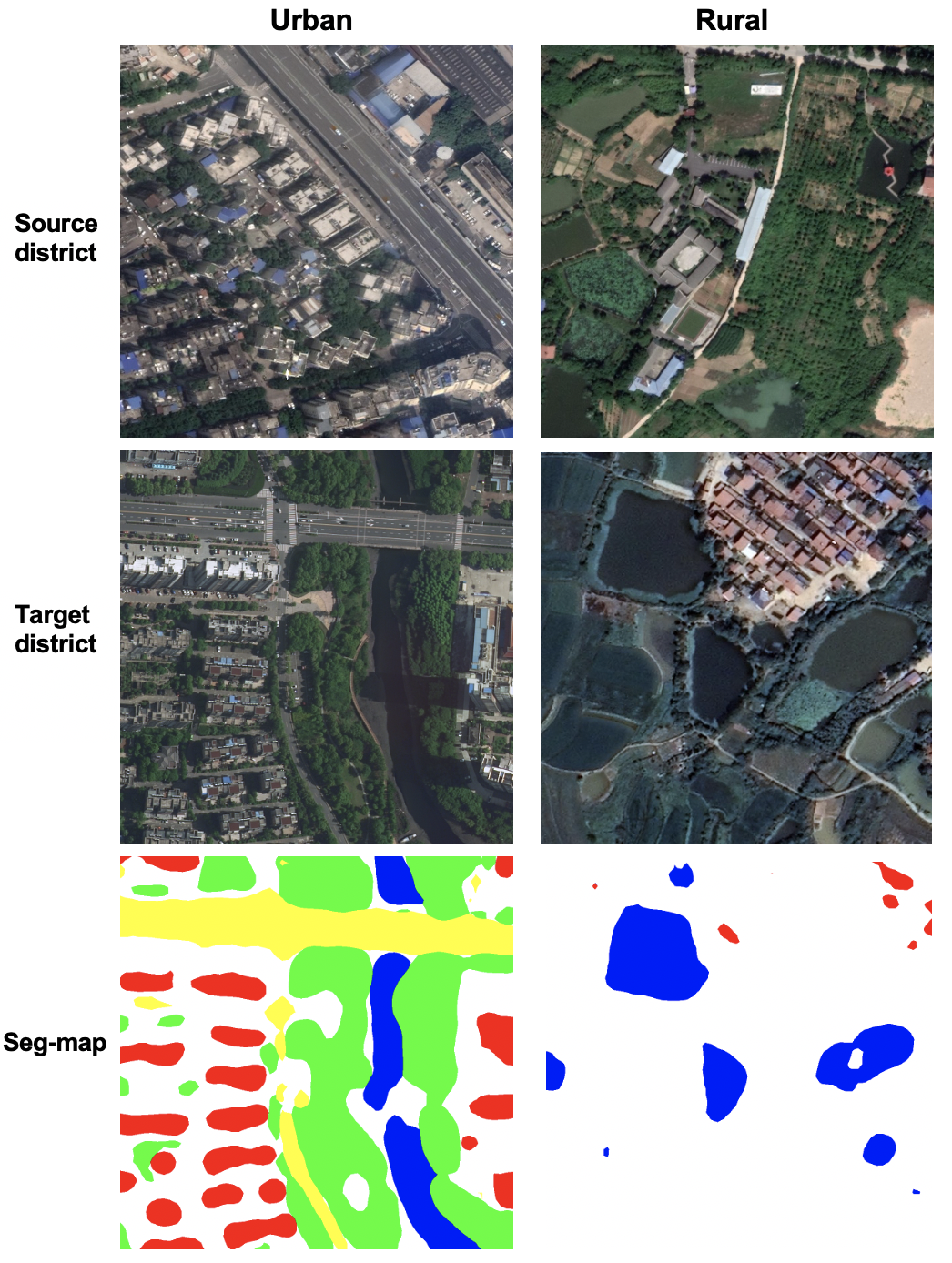}
    \caption{\textbf{Sample images from urban and rural scenes.} For each scene, one image from source domain districts, one from target domain districts, and the network's segmentation predictions (Seg-map) for the 7 land-cover classes on that target image are shown.}
    \label{fig:segmentation-map}
\end{figure}

\section{Dataset}

We use the publicly available, high spatial resolution land-cover dataset called LoveDA~\cite{wang2021loveda} in this study. Compared to other popular satellite image datasets, such as Zurich Summer ~\cite{volpi2015semantic}, DeepGlobe ~\cite{demir2018deepglobe}, and DSTL \cite{iglovikov2017dstl}, the recently released LoveDA has more annotated images and includes images from diverse locations. The dataset consists of 5987 images of size 1024$\times$1024 and spatial resolution 0.3m. The images are collected from 18 administrative districts from three cities in China, namely Nanjing, Changzhou, and Wuhan. Out of these, there are 9 urban districts and 9 rural districts, categorized based on their population density and level of economic development. We use satellite images from the 12 districts for which ground-truth masks are available: Gulou, Qinhuai, Qixia, Yuhuatai, Jintan, and Jianghan (urban); Pukou, Lishui, Liuhe, Huangpi, Gaochun, and Jiangxia (rural). The remaining 6 districts are not availanble as they are held out for the benchmark challenge.
The dataset contains segmentation masks, which are pixel-level labels, for 7 land-cover classes: Background, Building, Road, Water, Barren, Forest, and Agricultural. The Background class consists of any pixel not belonging to the other classes. Statistics of the dataset, given in Figure 3 in~\cite{wang2021loveda}, show that the pixel counts across the 7 classes are imbalanced. Further, distribution of classes and of building scales differ between urban and rural scenes. Thus the rural and urban groups of images, which we use in our fairness analysis, have different characteristics.

\section{Methods}

We study three tasks, namely semantic segmentation within the same domain, across districts, and across rural-urban areas. Next, we describe the setup for each task.

\subsection{Task A - Semantic segmentation}

Our task is to train multi-class semantic segmentation models for detecting the 7 land-cover classes from a given image. Sample images and predicted segmentation maps are shown in Figure \ref{fig:segmentation-map}. Same as the models studied in the LoveDA study~\cite{wang2021loveda}, we use two commonly used deep learning-based segmentation methods -- U-Net~\cite{ronneberger2015u} and DeepLabV3+~\cite{chen2018encoder} network, both with pre-trained ResNet50 ~\cite{he2016deep} as the backbone model for the encoder~\cite{garcia2018survey}. 
% based on the implementation in the original LoveDA study ~\cite{wang2021loveda}.

\subsubsection{Training-testing details}
Images from the 12 districts with labeled data are shuffled and split into training ($\approx$ 80\%) and testing sets ($\approx$ 20\%). Training set has 3148 images with a mix of 1377 urban and 1771 rural images, and the rest of the images comprise the testing set with a mix of 368 urban images and 473 rural images. Images are augmented during training by mirroring and rotation. Dimension of the input image to the network is 512$\times$512$\times$3 where 3 indicates the RGB bands. The output dimension of the network is 512$\times$512$\times$7, where 7 represents the probability of each pixel belonging to each land-cover class. We use cross-entropy (CE) loss, and stochastic gradient descent (SGD) as the optimizer with a momentum of 0.9 and a weight decay of $10^{-4}$. The batch size is set to 16 and the total training iterations are 15000, during which the learning rate is decayed using a polynomial learning rate scheduler implemented in PyTorch \cite{paszke2019pytorch}.

\begin{table*}[h!]
  \FloatBarrier
  \small
  \begin{center}
    \scalebox{0.9}{
    \begin{tabular}{*{6}{c|}c}
      \toprule 
        & \multicolumn{3}{c}{Mean} & \multicolumn{3}{|c}{Class-std} \\
      Model & \textbf{rural} & urban & \cellcolor{blue!15}{rural $-$ urban (\%)}  & rural & \textbf{urban}  & \cellcolor{blue!15}{rural $-$ urban (\%)} \\
      \midrule 
      UNet & \textbf{0.639} & 0.595 & \cellcolor{blue!15}{0.044 (6.9\%)} & 0.106 & 0.0946 & \cellcolor{blue!15}{0.0114 (10.8\%)} \\
      DeepLabV3+ & 0.632 & 0.597 & \cellcolor{blue!15}{0.035 (5.5\%)} & 0.0982 & \textbf{0.0896} & \cellcolor{blue!15}{0.0086 (8.8\%)}\\
      
      \midrule
      \midrule
      
      & \multicolumn{3}{c}{Worst} & \multicolumn{3}{|c}{Sorted 30\% (bottom, top) } \\
      Model & rural & \textbf{urban} & \cellcolor{blue!15}{rural $-$ urban (\%)}  & rural & urban  & \cellcolor{blue!15}{rural $-$ urban (\%)} \\
      \midrule 
      UNet & 0.453 & \textbf{0.474} & \cellcolor{blue!15}{$-$0.021 (-4.4\%)} & (0.491, 0.742) & (0.480, 0.705) & \cellcolor{blue!15}{(0.011 , 0.037)(2.2\%, 5.0\%)} \\
      DeepLabV3+ & 0.473 & 0.473 & \cellcolor{blue!15}{0 \quad (n/a)} & (0.504, 0.740) & (0.489, 0.706) & \cellcolor{blue!15}{(0.015, 0.034)(3.0\%, 4.6\%)} \\
      \bottomrule 
    \end{tabular}}
    \caption{\textbf{Task A: Evaluation on single-domain semantic segmentation.} Two networks, UNet and DeepLabV3+, are tested on rural and urban districts from the same domain as training set. For metrics, Mean, Class-std, and Worst, the better performing group (between rural and urban) is in bold. The difference in performance between rural and urban is shaded. Typically, performance is better for rural than urban.}
    \label{tab:table1}
  \end{center}
\end{table*}

\begin{table*}[h!]
  \small
  \begin{center}
    \scalebox{0.9}{
    \begin{tabular}{*{6}{c|}c}
      \toprule 
        & \multicolumn{3}{c}{Mean} & \multicolumn{3}{|c}{Class-std} \\
      Method & rural & \textbf{urban} & \cellcolor{blue!15}{rural $-$ urban (\%)} & rural & \textbf{urban} & \cellcolor{blue!15}{rural $-$ urban (\%)} \\
      \midrule 
      No adaptation & 0.364 & 0.486 & \cellcolor{blue!15}{$-$0.122 ($-$25\%)} & 0.200 & 0.135 & \cellcolor{blue!15}{0.065 (33\%)} \\
      CBST & 0.374 & \textbf{0.523} & \cellcolor{blue!15}{$-$0.149 ($-$28\%)} & 0.215 & \textbf{0.105} & \cellcolor{blue!15}{0.110 (51\%)} \\
      IAST & 0.376 & 0.493 & \cellcolor{blue!15}{$-$0.117($-$24\%)} & 0.223 & 0.135   & \cellcolor{blue!15}{0.088 (39\%)}\\

      \midrule
      \midrule
      
       & \multicolumn{3}{c}{Worst} & \multicolumn{3}{|c}{Sorted 30\% (bottom, top) } \\
       Method & rural & \textbf{urban} & \cellcolor{blue!15}{rural $-$ urban (\%)} & rural & urban & \cellcolor{blue!15}{rural $-$ urban (\%)} \\
       \midrule 
       No adaptation & 0.0609 & 0.244 & \cellcolor{blue!15}{$-$0.183 ($-$75\%)} & (0.098, 0.581) & (0.317, 0.630) & \cellcolor{blue!15}{($-$0.219, $-$0.049) ($-$69\%, $-$7.7\%)} \\
      CBST &  0.0172 & \textbf{0.362} & \cellcolor{blue!15}{$-$0.345 ($-$95\%)} & (0.0943, 0.609) & (0.398, 0.647) & \cellcolor{blue!15}{($-$0.304, $-$0.038) ($-$76\%, $-$5.9\%)}\\
      IAST &  0.0304 & 0.232 & \cellcolor{blue!15}{$-$0.202 ($-$87\%)} & (0.0772, 0.598) & (0.327, 0.640) & \cellcolor{blue!15}{($-$0.250, $-$0.042) ($-$76\%, $-$6.6\%)} \\
      
      \bottomrule 
       
    \end{tabular}}
    \caption{\textbf{Task B: Evaluation of transfer across districts.} Three methods (No adaptation, CBST, IAST) are trained source districts, and evaluated on target rural and target urban districts. For the metrics, Mean, Class-std, and Worst, the better performing group (between rural and urban) is in bold. The differences in performance between rural and urban are shaded. Models have high unfairness upon transfer.}
    \label{tab:table2}
  \end{center}
\end{table*}

\subsubsection{Evaluation metrics}
\label{sec:metrics}
% The test set includes 368 urban images and 473 rural images. 
We test the models on either the whole test set or the urban and rural subsets in the test set separately, and evaluate model performance using the following metrics:

\textbf{Accuracy metrics:} Intersection-over-Union (IoU), also called Jaccard index, is used to measure segmentation accuracy which is a common method to evaluate the quality of image segmentation~\cite{everingham2015pascal, wang2020image}. IoU for a class is defined as the intersection of class-wise ground-truth masks and the predicted segmentation divided by their union,

\begin{align*}
\text{IoU} := \frac{TP}{TP + FP + FN},
\end{align*}
where $TP$, $FP$, and $FN$ are pixel-wise true positives, false positives, and false negatives. We report IoU score of the model on each land-cover class as well as mean over class-wise IoU (referred to as \textbf{Mean}) over the 7 classes. 

\textbf{Fairness metrics:} Besides looking at IoU on the rural and urban subsets and comparing the two values, we devise three additional metrics that quantify how accuracy is distributed across the classes. The metrics have been used in existing fairness analysis of segmentation models \cite{szabo2021tilted}. These are:
\begin{enumerate}[leftmargin=*]
    \item \textbf{Class-std}. Standard deviation of IoUs across the 7 classes;
    \item \textbf{Worst}. IoU of the worst-performing class (Worst);
    \item \textbf{Sorted 30\% (bottom, top)}. Mean of the bottom 30\% and top 30\% classes of the sorted class IoUs. In our case, 30\% is 2 classes out of 7.
\end{enumerate}

Next, we describe the setup for the two transfer tasks.

\subsection{Transfer learning}

As mentioned earlier, we consider the setting of unsupervised domain adaption (UDA) that is we have a single task (image segmentation) on the two domains. We assume access to images and labels for the source (train) domain and only the images for the target (test) domain. This is a practical setting in satellite imagery since collecting images is inexpensive due to advancements in remote sensing, however, annotating the segmentation labels is expensive. Thus, we want to be able to use the labelled source images to segment known but as yet unlabelled target domain.

We consider two UDA methods which performed the best on the LoveDA benchmark~\cite{wang2021loveda} -- class-balanced self-training (\textbf{CBST}) ~\cite{zou2018unsupervised} and instance adaptive self-training (\textbf{IAST}) ~\cite{mei2020instance}. CBST optimizes the generation of pseudo-labels used during self-training to be more balanced among the classes by using class-wise confidence scores. IAST adaptively adjusts the pseudo-label generation to improve the diversity of pseudo-labels and saves useful information from hard instances.
We also use a natural method which ignores transfer learning and trains only with data from the source domain (\textbf{No adaptation}).

For all transfer learning experiments, we use a DeepLabV3+ ~\cite{chen2018encoder} network with ResNet50 encoder. An Adam optimizer is used with a momentum of 0.9. The batch size is 8 and the total training iterations are 15000. Other experimental setup parameters are the same as in the semantic segmentation task.

\subsubsection{Task B - Transfer across districts}
First, we consider the scenario where a model is transferred across different geographical locations, which in our case are administrative districts. \textbf{Source domain} comprises of 8 districts: Gulou, Qinhuai, Qixia, Jinghan (urban), and Pukou, Gaochun, Lishui, Jingxia (rural); and \textbf{Target domain} has 4 districts: Yuhuatai, Jintan (urban), and Liuhe, Huangpi (rural). The "No adaptation" and two UDA methods (CBST, IAST) are applied to train the network on the source domain, and are tested on urban and rural images from the unseen target domain, separately. The same accuracy and fairness metrics listed in Section \ref{sec:metrics} are used for evaluation.

\subsubsection{Task C - Transfer across urban and rural areas}
Second, we consider the scenario where the segmentation model is transferred either from urban to rural areas or from rural to urban areas. The source and target domain consists of data from the same set of districts. So for this task, the only source for domain discrepancy is rural and urban discrepancy. The no- adaptation method and two UDA methods are trained on the source domain, and tested on the target domain. Evaluation metrics are the same as earlier.

\begin{figure}[h!]
  \centering
    \includegraphics[width=8.3cm]{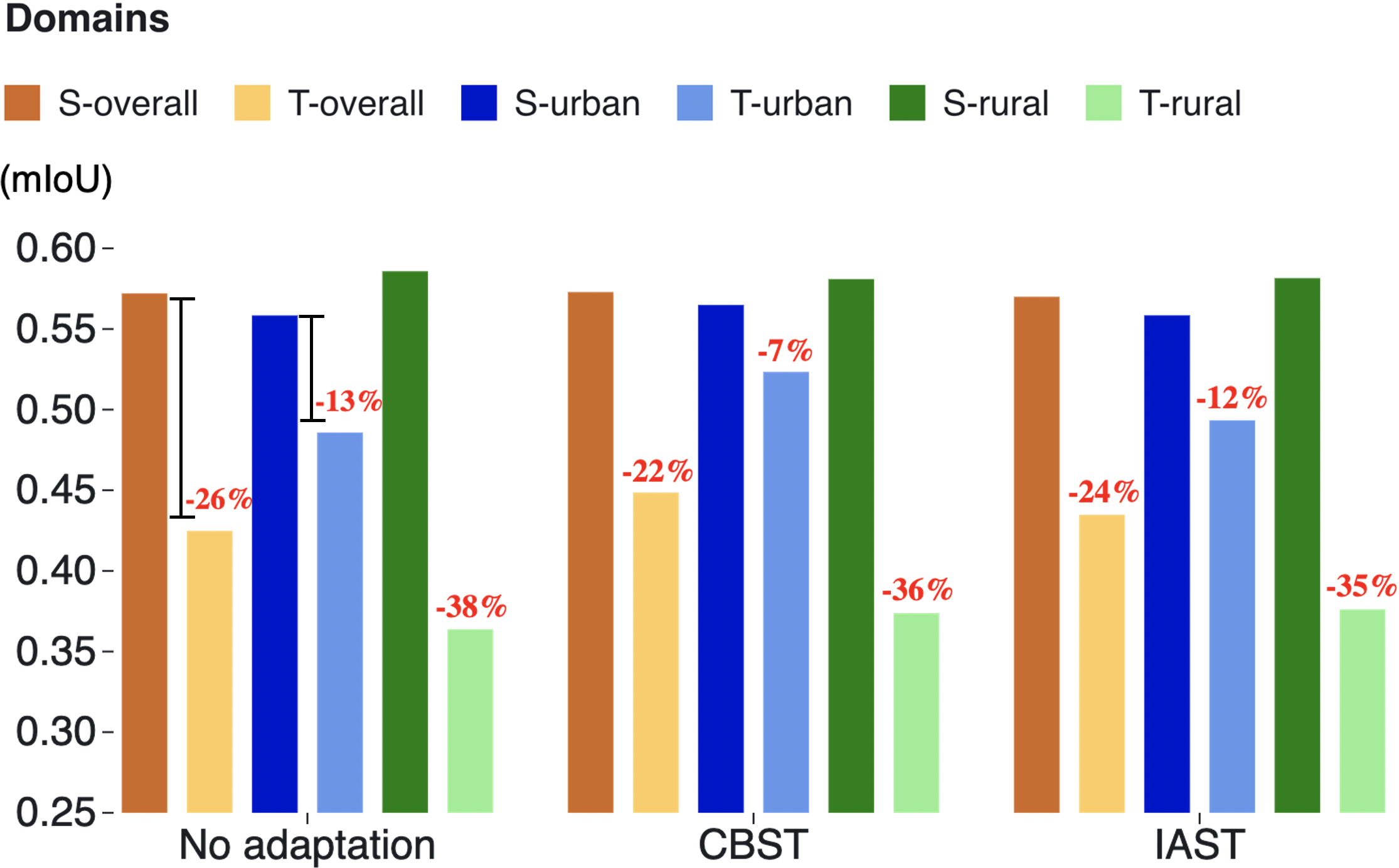}
    \caption{\textbf{Task B: Mean IoU upon transfer across districts.} Mean IoU on the union of rural and urban data from the source (S-overall) and target (T-overall), urban data from the source (S-urban) and target (T-urban), rural data from the source (S-rural) and target (T-rural) is plotted when transferring models across districts. No adaptation is the source-only method, CBST and IAST are UDA methods. While overall accuracy drops on transfer, UDA methods have smaller accuracy drop.}
    \label{fig:transfer-district}
\end{figure}

\begin{table*}[h!]
  \small
  \begin{center}
    \scalebox{0.9}{
    \begin{tabular}{*{5}{c|}c}
      \toprule 
      Sub-task & Method & Mean & Class-std & Worst & Sorted 30\% (bottom, top) \\
      \midrule 
       & No adaptation & 0.437 & 0.108 & 0.301 & (0.322, 0.566) \\
      Rural$\rightarrow$Urban & CBST & \textbf{0.469} & 0.123 & \textbf{0.326} & (\textbf{0.332}, 0.617) \\
      & IAST & 0.443 & 0.175  & 0.205 & (0.211, \textbf{0.638}) \\
      \midrule 
       & No adaptation & 0.426 & 0.108 & 0.226 & (0.271, 0.531) \\
      Urban$\rightarrow$Rural & CBST & \textbf{0.467} & 0.129 & 0.228 & (0.283, \textbf{0.599}) \\
      & IAST & 0.454 & 0.120  & \textbf{0.229} & (\textbf{0.307}, 0.592) \\
      \bottomrule 
    \end{tabular}}
    \caption{\textbf{Task C: Evaluation of urban-rural transfer.} Three methods (No adaptation, CBST, IAST) are trained on rural districts and evaluated on urban districts, and vice versa. Results with the most improvements are marked in bold. UDA methods improve Mean IoU compared to No adaptation but increase standard deviation of IoUs across classes.}
    \label{tab:table3}
  \end{center}
\end{table*}

\section{Results}

We summarize results for the single-domain in Table \ref{tab:table1}. Both the networks (UNet, DeepLabV3+) have better overall accuracy, shown with Mean IoU over the 7 classes, for the rural districts compared to the urban districts. Fairness metrics such as IoU of the worst class and mean IoU of 30\% bottom classes are comparable between rural and urban. The worst class is Barren for both rural and urban. The 30\% bottom classes include Barren and Road for rural, and Barren and Forest for urban (see Table ~\ref{appendix:table5} in Appendix for class-wise results). Rural results show higher mean IoU from top 30\% classes than urban results, but higher class-wise standard deviation. Overall, we observe rural-urban disparities in all four metrics.

For Task B on transfer learning across districts, we summarize the results in Figure \ref{fig:transfer-district} and Table \ref{tab:table2}. Figure \ref{fig:transfer-district} visualizes the mean IoU metric for both source and target districts. Based on the first two bar plots (dark and light orange) for No adapation, CBST, and IAST, we conclude that UDA methods improve overall segmentation accuracy on the target domain (T-overall) compared to No adaptation (a decrease of 22\% and 24\% vs that of 26\%). Similar trend is observed for each of the source-target pairs for rural and urban separately. 
However, the performance gap \textit{between} the rural and urban data from target (T-rural and T-urban) remains large. For instance, from Table \ref{tab:table2} we observe that CBST obtains mean IoU of 0.523 on urban area which is better than the "No adaptation" 0.486, and IAST obtains 0.376 on rural area better than the "No adaptation" 0.364. However, the networks remain unfair across rural-urban groups after the transfer (large values in the rural $-$ urban columns). UDA methods further lower fairness: CBST increases the difference of mean IoU between urban and rural by 22\% (-0.122 to -0.149), increases the difference of standard deviation by 69\% (0.065 to 0.11), and increases the difference of worst-performing class' IoU by 89\% (-0.183 to -0.345).
% The two UDA methods improve overall segmentation accuracy on target domain,

\begin{figure}[h!]
  \centering
    \includegraphics[width=8.3cm]{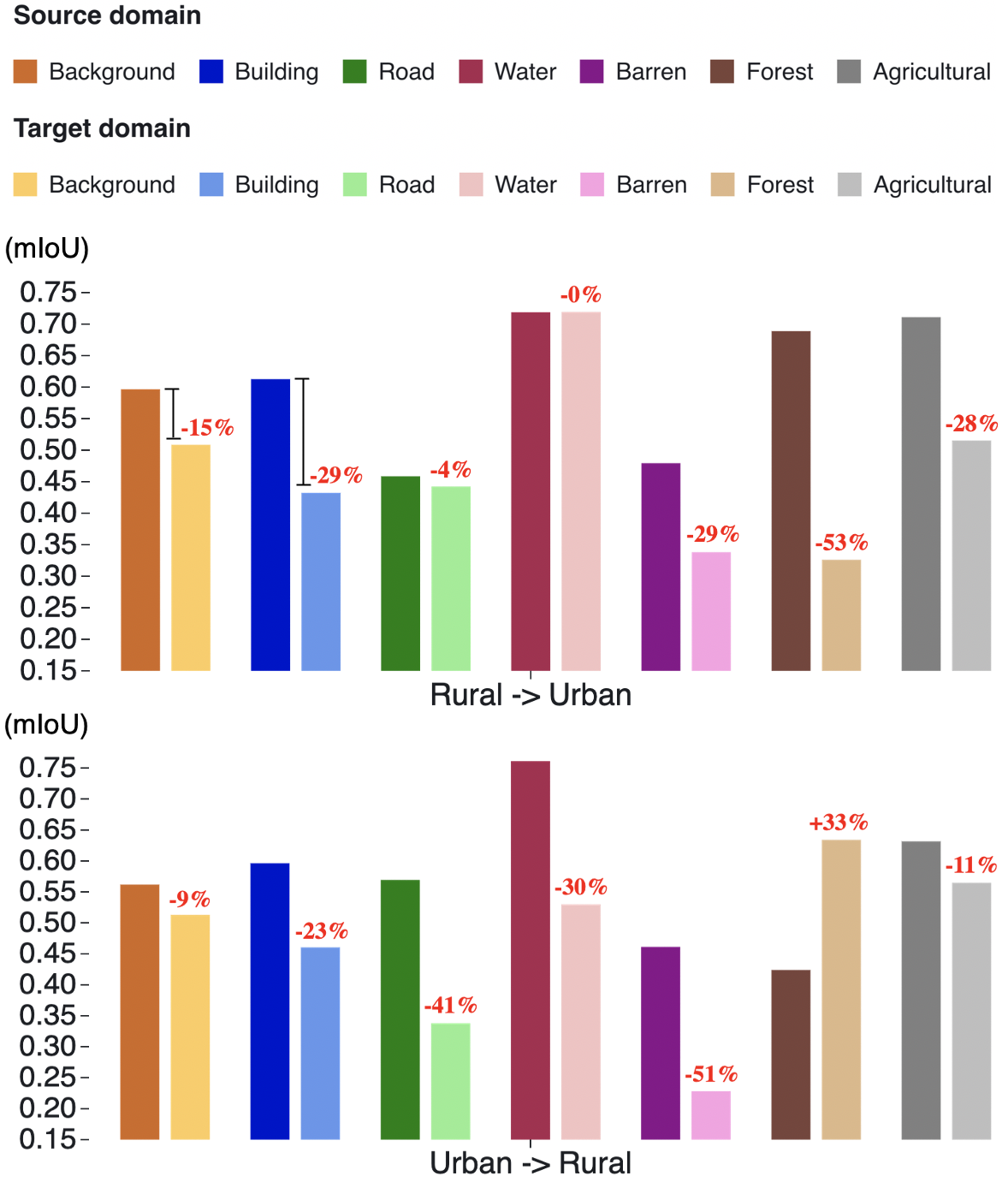}
    \caption{\textbf{Task C: Mean IoU upon transfer across rural-urban.} Mean IoU for 7 landscape classes on source and target domain when transferring from rural area to urban area, and from urban area to rural area, with the UDA method CBST. Performance changes vary substantially by class.}
    \label{fig:transfer-ru}
\end{figure}

Next, we examine Task C on transfer learning from urban domain to rural domain and vice versa. Results are summarized in Table \ref{tab:table3}. For both the transfer directions, the two UDA methods improve the overall accuracy, shown as higher mean IoU, higher IoU on the worst-preforming class, and higher mean IoU on bottom and top 30\% classes. However, compared to "No adaptation", UDA methods disperse model performance across the classes, measured by higher standard deviation. For example, CBST increases Class-std from 0.108 to 0.123 on rural to urban transfer and from 0.108 to 0.129 on urban to rural transfer. Looking more closely into the CBST method which obtains the best overall transfer accuracy (0.469 and 0.467), its performance on each class is visualized in Figure \ref{fig:transfer-ru}. We observe that the IoU changes for each class upon transfer are highly unequal. For example, in transferring from Rural$\rightarrow$Urban, the network retains accuracy on the Water and Road classes, but lost significant accuracy ($53\%$) on the Forest class. In transferring from Urban$\rightarrow$Rural, accuracy drops significantly on Road, Water, and Barren classes, but increases by 33\% on the Forest class.

% In summary, while transfer learning helps in overall accuracy on the target domain, it increases the disparity in performance across rural-urban locations or across the classes.

\section{Discussion}
For the locations included in this study, segmentation results showed a disparity in performance between rural and urban areas. Though the two groups obtain similar accuracy on the respective bottom 30\% performing land-cover classes, rural areas obtain better accuracy on the top 30\% performing classes. Moreover, performance distribution across classes are different between rural and urban images. Specifically, the segmentation model detects Forest and Agriculture classes well in their rural form, and detects Road and Water classes well in their urban form (detailed results are reported in Table \ref{tab:table5} in the Appendix). Due to urbanization, rural and urban areas have clear landscape differences. For example, roads are typically wider in the urban scenes and narrower in rural scenes and water takes on larger shapes like lakes in urban scenes, and smaller shapes like ditches in rural scenes~\cite{wang2021loveda}. This may explain why the networks show advantages in urban images on Road and Water classes. Moreover, agricultural land covers large area and is continuously distributed in rural scenes. The percentage of pixels with Agriculture and Forest elements is also higher~\cite{wang2021loveda}. This can facilitate learning on these two classes in rural areas as compared to urban areas.

We considered two practical transfer learning tasks with satellite images and assess network fairness while transferring across geographical locations, and across rural and urban areas. Broadly, we observed that when transferring across districts, networks made more unfair predictions on data from the new domain than data from the same domain as training. For the network trained without any adaptation, the mean IoU accuracy difference between rural and urban images on the target domain is around 64\% higher than the difference reported in the single-domain task. Similarly, all other fairness metrics show much higher differences between groups on transferring to the target domain. Notably, when applying UDA methods, CBST and IAST, transfer accuracy was improved, but at the cost of fairness damage. These methods further enlarged the performance gap between rural and urban groups measured in all four metrics. These findings indicate a need for new domain adaptation methods that tackle the challenge of fair transfer learning.

\begin{table}[h!]
  \small
  \begin{center}
    \scalebox{0.9}{
    \begin{tabular}{*{4}{c|}c}
      \toprule 
      Group & Source & Target & PAD & MMD  \\
      \midrule 
      Urban & Gulou & Yuhuatai & 0.26 & 0.207 \\
       & Qinhuai & Jintan & ($\pm$0.12) &  ($\pm$0.0235)  \\
       & Qixia &  &  &   \\
       & Jinghan &  &  &   \\
     
      \midrule 
      Rural & Pukou & Liuhe & 0.64 & 0.262  \\
       & Gaochun & Huangpi & ($\pm$0.21) & ($\pm$0.0437) \\
       & Lishui &  &  &   \\
       & Jingxia &  &  &  \\
      \bottomrule 
    \end{tabular}}
    \caption{\textbf{Shift in raw image distribution.} Measurement of source-target domain distance using two metrics -- Proxy-A-distance (PAD) and Maximum mean discrepancy (MMD). The implementation is based on the online codebase in ~\cite{transferlearning.xyz}. Rural images shift more than urban images.}
    \label{tab:table4}
  \end{center}
\end{table}

One of the possible reasons why the network can better adapt to urban images than rural images is the unequal domain discrepancy. To estimate how similar the source and target images are in the rural and urban groups, we use two metrics -- Proxy-A-distance (PAD) ~\cite{ganin2016domain} and Maximum mean discrepancy (MMD) ~\cite{gretton2012kernel}. Both measure the dissimilarity between data distributions of different domains. We randomly sample 100 images from each domain at a time and ran 30 trials to compute the two measures. The mean and standard deviation of distance across the trials are reported in Table ~\ref{tab:table4}. We observe that the raw satellite images are overall more dissimilar between source and target districts for rural than for urban, which is a likely cause of the unequal transfer learning performance between the two groups. Figure ~\ref{fig:segmentation-map} illustrates example images from both domains and the segmentation predictions our network made on the target. We see that Buildings across source and target districts are of similar shape and arrangement for urban, but they are disordered and dissimilar for rural. Accordingly, the model segmentation map shows that the model segments urban buildings well but fails to detect most of the rural buildings. This observation indicates the importance of checking class differences besides overall differences across the whole image. 

\begin{figure}
  \centering
    \includegraphics[width=8.2cm]{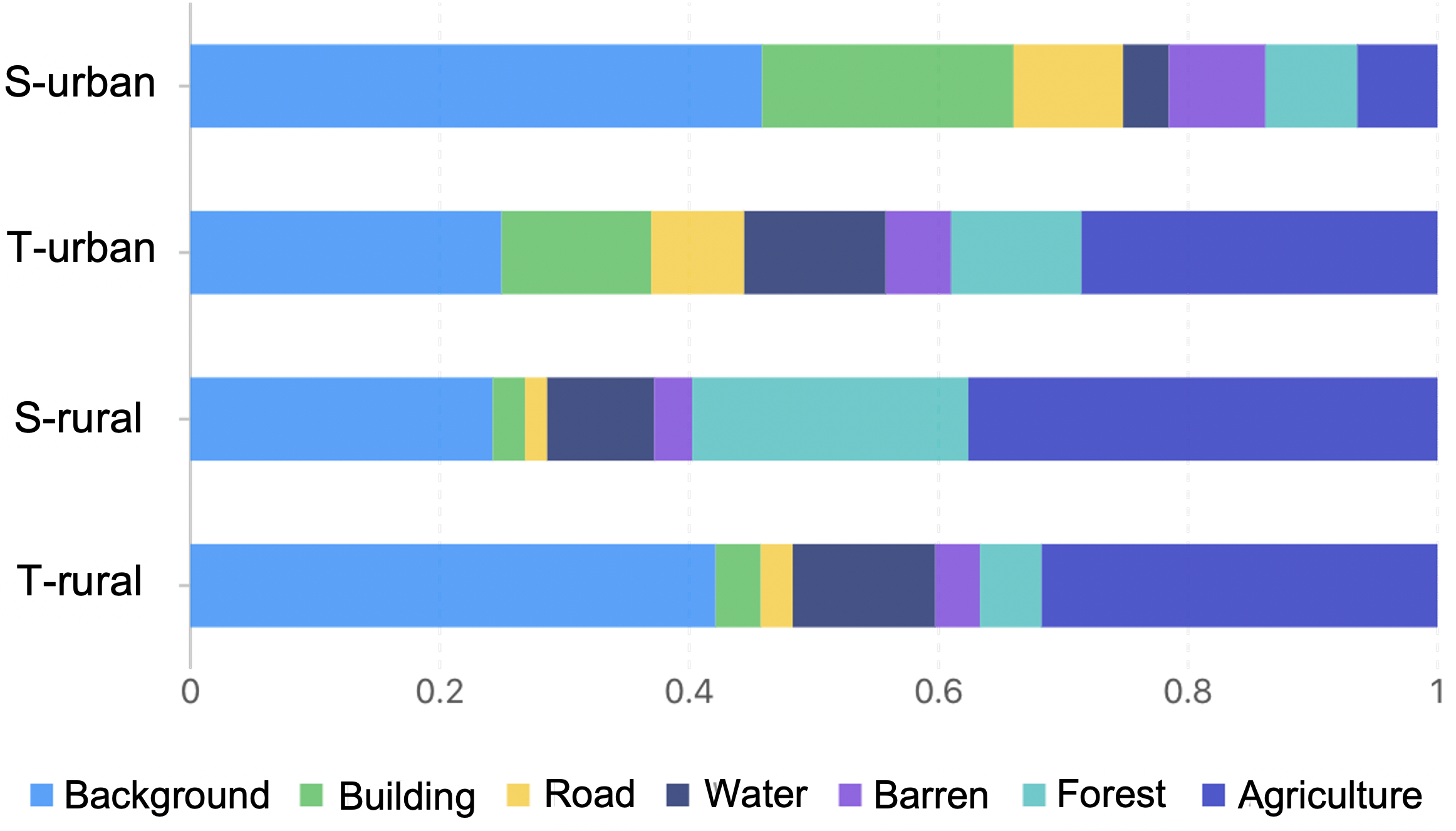}
    \caption{\textbf{Shifts in class distribution.} Class distribution in terms of proportion of pixels per class is plotted for urban images from source domain (S-urban), urban images from target domain (T-urban), rural images from source domain (S-rural), and rural images from target domain (T-rural). Class distribution is substantially different for all subsets.}
    \label{fig:pixels}
\end{figure}

Along these lines, we define pixel-wise class distribution as the proportion of pixels belonging to each class. We assess shifts in the class distribution between source and target for both rural and urban images. Class distributions are shown in Figure ~\ref{fig:pixels}. For urban locations, the Water and Agriculture classes have the largest shifts from source to target. For rural, the Background and Forest classes have the largest shifts. Indeed there are large class shifts between source and target for both the urban and rural data. For example, the class distribution of urban target data seems more rural-like with an increased proportion in the Agriculture class. This emphasizes the internal variation in rural and urban categories. Moreover, as our data consists of images from just one set of locations, data from different locations are needed for more generalizable conclusions. However, based on the selection of classes examined in our data which are common land-use classes globally, some results (such as in Table \ref{tab:table4} indicate common threads that can be applicable to rural-urban disparities in general.

In the second transfer learning task, the networks show unfairness on rural to urban domain transfer. Differences between rural and urban scenes provide explanation for why almost all classes lost accuracy on the target domain. Some classes show opposite transfer performance in the two sub-tasks of Task C. The Road class lost only 4\% accuracy on Rural$\rightarrow$Urban transfer but lost 41\% accuracy on Urban$\rightarrow$Rural transfer. The Water class shows similar patterns, whereas the Forest class lost 53\% accuracy on Rural$\rightarrow$Urban transfer but gains 33\% accuracy on Urban$\rightarrow$Rural transfer. These observations indicate that for some classes, the features learnt by the networks from rural scenes can be easily adapted to interpret urban scenes, and some classes have the opposite case. From this perspective, features of different classes can have very different generalization ability, which will cause the transfer performance to be highly unequal across classes. This feature-specific characteristic may be leveraged in the design of future transfer learning methods.

\section{Conclusion}
Transfer learning models for semantic segmentation are often evaluated based on overall accuracy metrics. Here, we expand the scope of their evaluation by conducting a systematic fairness evaluation when models are transferred across domains. We examine the performance of two unsupervised domain adaptation methods on a large-scale public satellite imagery dataset. Model fairness is evaluated between rural and urban locations as the models are trained and tested across administrative districts. Based on the experiments, we conclude that the domain adaptation methods we study can be improved in terms of retaining model fairness across rural and urban data. Domain adaptation improves overall accuracy at the cost of decreasing fairness on test domain. Further, more shifts in the raw image distribution and pixel-wise class distribution result in more performance drop. Broadly, our findings demonstrate potential fairness problems when working with satellite image data sourced from different locations. Also, the findings indicate the need for developing methods focused on fair transfer learning, such as through new model architectures or loss functions.
% based on single domain or multiple domain via transfer learning, without the fairness issues.
%%%%%%%%% REFERENCES
{\small
\bibliographystyle{ieee_fullname}
% \bibliography{egbib}

}

\clearpage
\appendix

\section*{Appendix}

\vspace{-1mm}
\section{Semantic segmentation results on single classes}
\label{appendix:details}
\vspace{-1mm}

In this section, we provide complete IoU result of the model on each land-cover class, for the semantic segmentation task and two transfer learning tasks in experiments.

\subsection{IoU on each land-cover class in the single-domain satellite image semantic segmentation task}
\label{appendix:table5}
\begin{table}[h!]
  \small
  \begin{center}
    \begin{tabular}{*{4}{c|}c}
      \toprule 
       & \multicolumn{2}{c}{UNet} & \multicolumn{2}{c}{DeepLabV3+} \\
      Class & Rural & Urban & Rural & Urban \\
      
      \midrule 
      Background & 0.6044 & 0.5843 & 0.6126 & 0.5912\\
      Building & 0.6678 & 0.6188 & 0.5965 & 0.5912 \\
      Road & 0.5297 & 0.5937 & 0.5341 & 0.6021 \\
      Water & 0.7501 & 0.7801 & 0.7453 & 0.7711 \\
      Barren & 0.4526 & 0.4742 & 0.4732 & 0.4729 \\
      Forest & 0.7345 & 0.4859 & 0.7356 & 0.5049 \\
      Agriculture & 0.7324 & 0.6291 & 0.7194 & 0.6411 \\
      \bottomrule 
    \end{tabular}
    \caption{Intersection over Union (IoU) of network UNet and DeepLabv3 on land-cover classes on testing images from rural districts and urban districts.}
    \label{tab:table5}
  \end{center}
\end{table}

\subsection{IoU on each land-cover class in the across-districts transfer learning task}

\label{appendix:table6}
\FloatBarrier
\begin{table}[h!]
  \small
  \begin{center}
    \begin{tabular}{*{4}{c|}c}
      \toprule 
       & \multicolumn{2}{c}{Rural} & \multicolumn{2}{c}{Urban} \\
      Class & Source & Target & Source & Target \\
      
      \midrule 
      Background & 0.4981 & 0.5517 & 0.6324 & 0.3898 \\
      Building & 0.4774 & 0.4536 & 0.5392 & 0.5760 \\
      Road & 0.4445 & 0.2440 & 0.5749 & 0.5596 \\
      Water & 0.7022 & 0.6105 & 0.7217 & 0.6835 \\
      Barren & 0.4674 & 0.1352 & 0.4761 & 0.2437 \\
      Forest & 0.7503 & 0.0609 & 0.4876 & 0.5175 \\
      Agriculture & 0.7607 & 0.4907 & 0.4767 & 0.4305 \\
      \bottomrule 
    \end{tabular}
    \caption{Intersection over Union (IoU) of network DeepLabv3, using \textbf{No adaptation}, on each land-cover class on testing images from source domain and target domain when transferring model across geographical locations,  evaluated on rural districts and urban districts, seperately.}
    \label{tab:table6}
  \end{center}
\end{table}

\label{appendix:table7}
\FloatBarrier
\begin{table}[h!]
  \small
  \begin{center}
    \begin{tabular}{*{4}{c|}c}
      \toprule 
       & \multicolumn{2}{c}{Rural} & \multicolumn{2}{c}{Urban} \\
      Class & Source & Target & Source & Target \\
      
      \midrule 
      Background & 0.4794 & 0.5823 & 0.6397 & 0.4339 \\
      Building & 0.4687 & 0.4042 & 0.5055 & 0.5741 \\
      Road & 0.4504 & 0.2593 & 0.5846 & 0.5484 \\
      Water & 0.7004 & 0.6313 & 0.7703 & 0.7188 \\
      Barren & 0.4775 & 0.1713 & 0.4867 & 0.3620 \\
      Forest & 0.7453 & 0.01723 & 0.4333 & 0.4858 \\
      Agriculture & 0.7449 & 0.5509 & 0.5340 & 0.5406 \\
      \bottomrule 
    \end{tabular}
    \caption{Intersection over Union (IoU) of network DeepLabv3, using UDA method \textbf{CBST}, on each land-cover class on testing images from source domain and target domain when transferring model across geographical locations,  evaluated on rural districts and urban districts, seperately.}
    \label{tab:table7}
  \end{center}
\end{table}

\label{appendix:table8}
\FloatBarrier
\begin{table}[h!]
  \small
  \begin{center}
    \begin{tabular}{*{4}{c|}c}
      \toprule 
       & \multicolumn{2}{c}{Rural} & \multicolumn{2}{c}{Urban} \\
      Class & Source & Target & Source & Target \\
      
      \midrule 
      Background & 0.4767 & 0.5850 & 0.6417 & 0.4218 \\
      Building & 0.5004 & 0.4930 & 0.4766 & 0.5702 \\
      Road & 0.4353 & 0.2340 & 0.5545 & 0.5280 \\
      Water & 0.6939 & 0.6107 & 0.7657 & 0.7101 \\
      Barren & 0.4627 & 0.1240 & 0.4878 & 0.2319 \\
      Forest & 0.7505 & 0.0304 & 0.4580 & 0.4778 \\
      Agriculture & 0.7514 & 0.5556 & 0.5247 & 0.5138 \\
      \bottomrule 
    \end{tabular}
    \caption{Intersection over Union (IoU) of network DeepLabv3, using UDA method \textbf{IAST}, on each land-cover class on testing images from source domain and target domain when transferring model across geographical locations,  evaluated on rural districts and urban districts, seperately.}
    \label{tab:table8}
  \end{center}
\end{table}

\subsection{IoU on each land-cover class in the crossing rural-urban transfer learning task}

\label{appendix:table9}
\FloatBarrier
\begin{table}[h!]
  \small
  \begin{center}
    \begin{tabular}{*{4}{c|}c}
      \toprule 
       & \multicolumn{2}{c}{Rural $\rightarrow$ Urban} & \multicolumn{2}{c}{Urban $\rightarrow$ Rural} \\
      Class & Source & Target & Source & Target \\
      
      \midrule 
      Background & 0.6172 & 0.4681 & 0.5976 & 0.4989  \\
      Building & 0.5868 & 0.4220 & 0.5639 & 0.4153   \\
      Road & 0.4499 & 0.4082 & 0.5793 & 0.3148   \\
      Water & 0.7199 & 0.6643 & 0.7454 & 0.4634  \\
      Barren & 0.4566 & 0.3008 & 0.5175 & 0.2264   \\
      Forest & 0.7066 & 0.3431 & 0.4752 & 0.5337   \\
      Agriculture & 0.7495 & 0.4491 & 0.6366 & 0.5277   \\
      \bottomrule 
    \end{tabular}
    \caption{Intersection over Union (IoU) of network DeepLabv3, using \textbf{No adaptation}, on each land-cover class on testing images from source domain and target domain when transferring model Rural $\rightarrow$ Urban and Urban $\rightarrow$ Rural.}
    \label{tab:table9}
  \end{center}
\end{table}

\label{appendix:table10}
\begin{table}[h!]
  \small
  \begin{center}
    \begin{tabular}{*{4}{c|}c}
      \toprule 
       & \multicolumn{2}{c}{Rural $\rightarrow$ Urban} & \multicolumn{2}{c}{Urban $\rightarrow$ Rural} \\
      Class & Source & Target & Source & Target \\
      
      \midrule 
      Background & 0.5968 & 0.5085 & 0.5618 & 0.5130  \\
      Building & 0.6126 & 0.4324 & 0.5963 & 0.4605   \\
      Road & 0.4585 & 0.4424 & 0.5692 & 0.3380   \\
      Water & 0.7190 & 0.7193 & 0.7611 & 0.5301  \\
      Barren & 0.4793 & 0.3381 & 0.4613 & 0.2281   \\
      Forest & 0.6890 & 0.3260 & 0.4239 & 0.6339   \\
      Agriculture & 0.7109 & 0.5152 & 0.6315 & 0.5646   \\
      \bottomrule 
    \end{tabular}
    \caption{Intersection over Union (IoU) of network DeepLabv3, using UDA method \textbf{CBST}, on each land-cover class on testing images from source domain and target domain when transferring model Rural $\rightarrow$ Urban and Urban $\rightarrow$ Rural.}
    \label{tab:table10}
  \end{center}
\end{table}

\label{appendix:table11}
\begin{table}[h!]
  \small
  \begin{center}
    \begin{tabular}{*{4}{c|}c}
      \toprule 
       & \multicolumn{2}{c}{Rural $\rightarrow$ Urban} & \multicolumn{2}{c}{Urban $\rightarrow$ Rural} \\
      Class & Source & Target & Source & Target \\
      
      \midrule 
      Background & 0.5968 & 0.5085 & 0.5618 & 0.5130  \\
      Building & 0.6126 & 0.4324 & 0.5963 & 0.4605   \\
      Road & 0.4585 & 0.4424 & 0.5692 & 0.3380   \\
      Water & 0.7190 & 0.7193 & 0.7611 & 0.5301  \\
      Barren & 0.4793 & 0.3381 & 0.4613 & 0.2281   \\
      Forest & 0.6890 & 0.3260 & 0.4239 & 0.6339   \\
      Agriculture & 0.7109 & 0.5152 & 0.6315 & 0.5646   \\
      \bottomrule 
    \end{tabular}
    \caption{Intersection over Union (IoU) of network DeepLabv3, using UDA method \textbf{IAST}, on each land-cover class on testing images from source domain and target domain when transferring model Rural $\rightarrow$ Urban and Urban $\rightarrow$ Rural.}
    \label{tab:table11}
  \end{center}
\end{table}

\end{document}